\documentclass[10pt,twocolumn,letterpaper]{article}

\usepackage{iccv}
\usepackage{times}
\usepackage{epsfig}
\usepackage{graphicx}
\usepackage{amsmath}
\usepackage{amssymb}

\usepackage{url}            %
\usepackage{booktabs}       %
\usepackage{amsfonts}       %
\usepackage{nicefrac}       %
\usepackage{microtype}      %
\usepackage{csquotes}
\usepackage{array}
\usepackage{adjustbox}
\usepackage{xspace}
\usepackage{multirow}
\usepackage{wrapfig}
\usepackage{import}

\usepackage{diagbox}
\usepackage{caption}
\usepackage{subcaption}
\usepackage{makecell}
\usepackage{enumitem}
\usepackage{gensymb}
\usepackage{amsmath,amssymb}
\usepackage{wrapfig}
\usepackage{ragged2e}
\usepackage{algorithm,algpseudocode}
\usepackage{color, colortbl}
\usepackage{bbm}

\definecolor{Gray}{gray}{0.85}
\DeclareSymbolFont{largesymbolsCM}{OMX}{cmex}{m}{n}
\DeclareMathSymbol{\sumop}{\mathop}{largesymbolsCM}{"50}

\usepackage[pagebackref=true,breaklinks=true,colorlinks,bookmarks=false]{hyperref}

\def\ours{\texttt{\textbf{VideoIQ}}\xspace}
\DeclareMathOperator*{\argmax}{arg\,max}

\usepackage{hyperref}

\iccvfinalcopy %

\def\iccvPaperID{6281} %

\ificcvfinal\fi

\begin{document}

\title{Dynamic Network Quantization for Efficient Video Inference}

\author{Ximeng Sun$^{1}$ \ \ \ \ \ \ \ \ \ \ \ Rameswar Panda$^{2}$ \ \ \ \ \ \ \ \ \ \ \ Chun-Fu (Richard) Chen$^{2}$ \\  Aude Oliva$^{2,3}$ \ \ \ \ \ \ \ \ \ \ \  Rogerio Feris$^{2}$  \ \ \ \ \ \ \ \ \ \ \ Kate Saenko$^{1, 2}$ \\
 $^{1}$Boston University, $^{2}$MIT-IBM Watson AI Lab, $^{3}$MIT 
}

\maketitle
\thispagestyle{empty} 

\begin{abstract}

Deep convolutional networks have recently achieved great success in video recognition, yet their practical realization remains a challenge due to the large amount of computational resources required to achieve robust recognition. Motivated by the effectiveness of quantization for boosting efficiency, in this paper, we propose a dynamic network quantization framework, that selects optimal precision for each frame conditioned on the input for efficient video recognition. Specifically, given a video clip, we train a very lightweight network in parallel with the recognition network, to produce a dynamic policy indicating which numerical precision to be used per frame in recognizing videos. We train both networks effectively using standard backpropagation with a loss to achieve both competitive performance and resource efficiency required for video recognition. Extensive experiments on four challenging diverse benchmark datasets demonstrate that our proposed approach provides significant savings in computation and memory usage while outperforming the existing state-of-the-art methods. Project page: \url{https://cs-people.bu.edu/sunxm/VideoIQ/project.html}.

\end{abstract}

\section{Introduction} \label{sec:introduction}

With the availability of large-scale video datasets~\cite{carreira2017quo,monfort2018moments}, deep learning models based on 2D/3D convolutional neural networks (CNNs)~\cite{chen2020deep,wang2016temporal,tran2015learning,karpathy2014large,hara2018can} have dominated the field of video recognition. However, despite impressive performance on standard benchmarks, efficiency remains a great challenge for many resource constrained applications due to the heavy computational burden of deep CNN models.

\begin{figure}
\begin{center}
     \includegraphics[width=\linewidth]{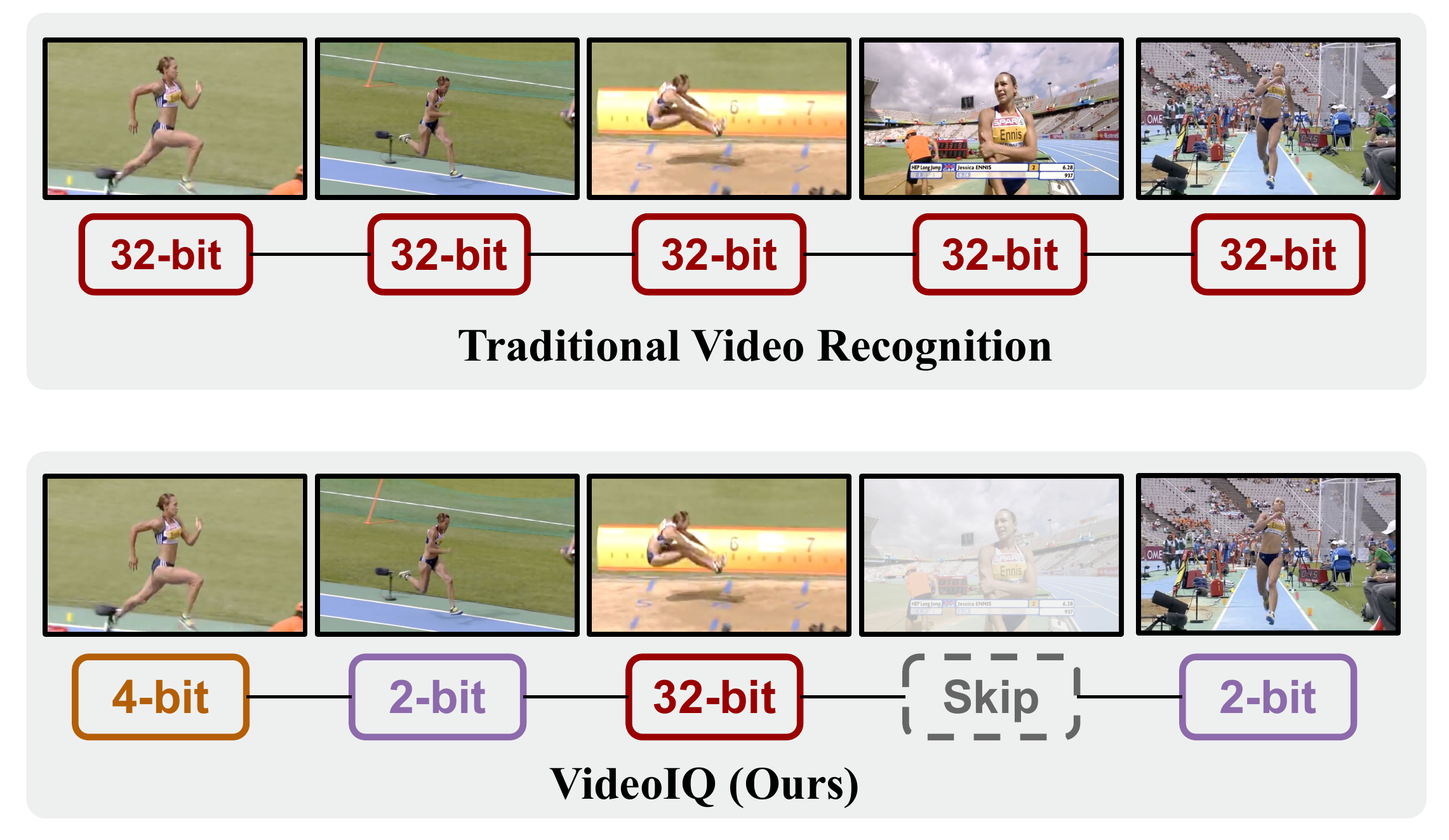}
\end{center} \vspace{-15pt}
   \caption{\small \textbf{A conceptual overview of our approach.} Instead of processing all the video frames with the same 32-bit precision, \ours learns to dynamically select optimal quantization precision conditioned on input clips for efficient video recognition. It is computationally very efficient to process more informative frames with high precision and less informative ones with lower precision, without sacrificing accuracy. Best viewed in color.
   }
   \vspace{-7pt}
   \label{fig:concept}
\end{figure}

Motivated by the need of efficiency, existing research efforts mainly focus on either designing compact models~\cite{piergiovanni2019tiny,tran2019video,feichtenhofer2020x3d} or sampling of salient frames for efficient recognition~\cite{wu2019adaframe,wu2019multi,meng2020ar}. While these methods have shown promising results, they all use 32-bit precision for processing all the frames in a given video, limiting their achievable efficiency. 
Specifically, orthogonal to the network design, the computational cost of a CNN is directly affected by the bit-width of weights and activations~\cite{han2015deep,zhou2016dorefa,choi2018pact}, which surprisingly as another degree of freedom for efficient video inference, is almost overlooked in previous works. To illustrate this, let us consider the video in Figure~\ref{fig:concept}, represented by five uniformly sampled frames. A quick glance on the video clearly shows that only the third frame can be processed using 32-bit precision as this is the most informative frame for recognizing the action ``Long Jump'', while the rest can be processed at very low precision or even skipped (i.e., precision set to zero) without sacrificing the accuracy (Bottom), resulting in large computational savings compared to processing all frames with same 32-bit precision, as generally done in mainstream video recognition methods (Top). 

Inspired by this observation, we introduce \textbf{Video} \textbf{I}nstance-aware \textbf{Q}uantization (\ours), which for the first time advocates a novel input-dependent dynamic network quantization strategy for efficient video recognition. While dynamic network quantization looks trivial and handy at the first glance, we need to address two challenges: (1) how to efficiently determine what quantization precision to use per target instance; and (2) given instance-specific precisions, how can we flexibly quantize the weights and activations of a single deep recognition network into various precision levels, without additional storage or computation cost.

To address the aforementioned challenges, we propose a simple end-to-end differentiable approach to learn a decision policy that selects optimal precision conditioned on the input, while taking both accuracy and efficiency into account in recognizing complex actions. We achieve this by sampling the policy from a discrete distribution parameterized by the output of a lightweight policy network, which decides on-the-fly what precision should be used on a per frame basis. Since these decision functions are discrete and non-differentiable, we train the policy network using standard back-propagation through Gumbel Softmax sampling~\cite{jang2016categorical}, without resorting to complex reinforcement learning, as in~\cite{wu2019adaframe,fan2018watching,yeung2016end}. Moreover, instead of storing separate precision-specific models, we train a single deep neural network for action recognition using joint training, 
which enables us to directly adjust the numerical precision by simply truncating the least significant bits, without performance degradation. Our proposed approach provides not only high computational efficiency but also significant savings in memory--a practical requirement of many real-world applications which has been largely ignored by prior works~\cite{meng2020ar,wu2019liteeval,meng2021adafuse,wu2019adaframe}.

We conduct extensive experiments on four standard video recognition datasets (ActivityNet-v1.3~\cite{caba2015activitynet}, FCVID~\cite{jiang2017exploiting}, Mini-Sports1M~\cite{karpathy2014large} and Mini-Kinetics~\cite{carreira2017quo}) to demonstrate the superiority of our proposed approach over state-of-the-art methods. Our results show that \ours can yield significant savings in computation and memory (e.g., average $26.0\%$ less GFLOPS and $55.8\%$ less memory), while achieving better recognition performance, over the most competitive SOTA baseline~\cite{meng2020ar}. We also discover that the decision policies learned using our method are transferable to unseen classes and videos across different datasets. Furthermore, qualitative results suggest that our learned policies correlate with the distinct visual patterns in video frames, i.e., our method utilizes 32-bit full precision only for relevant video frames and process non-informative frames at low precision or skip them for computation efficiency.

\section{Related Work} \label{sec:relatedwork} 

\vspace{1mm}
\noindent\textbf{Video Recognition.} Much progress has been made in developing a variety of ways to recognize videos, by either applying 2D-CNNs~\cite{karpathy2014large,wang2016temporal,Simonyan14TwoStream,sudhakaran2020gate} or 3D-CNNs~\cite{tran2015learning,carreira2017quo,hara2018can}. Despite promising results, there is a significant interest in developing more efficient models with reasonable performance~\cite{piergiovanni2019tiny,tran2019video}. 
SlowFast network~\cite{feichtenhofer2019slowfast} employs two pathways for recognizing actions by processing a video at both slow and fast frame rates. 
Many works utilize 2D-CNNs for efficient recognition by modeling temporal causality using different aggregation modules~\cite{wang2016temporal,zhou2018temporal,fan2019more,lin2019tsm}. 
Expansion of 2D architectures across frame rate, spatial resolution, network width, is proposed in~\cite{feichtenhofer2020x3d}. 
While these approaches bring reasonable efficiency improvements, all of them process the video frames using same 32-bit precision, regardless of information content in each input frame, which varies in most real-world long videos. In contrast, our approach dynamically selects bit-width per input, to strategically allocate computation at test time for efficient recognition. 

\begin{figure*}
\begin{center}
     \includegraphics[width=\linewidth]{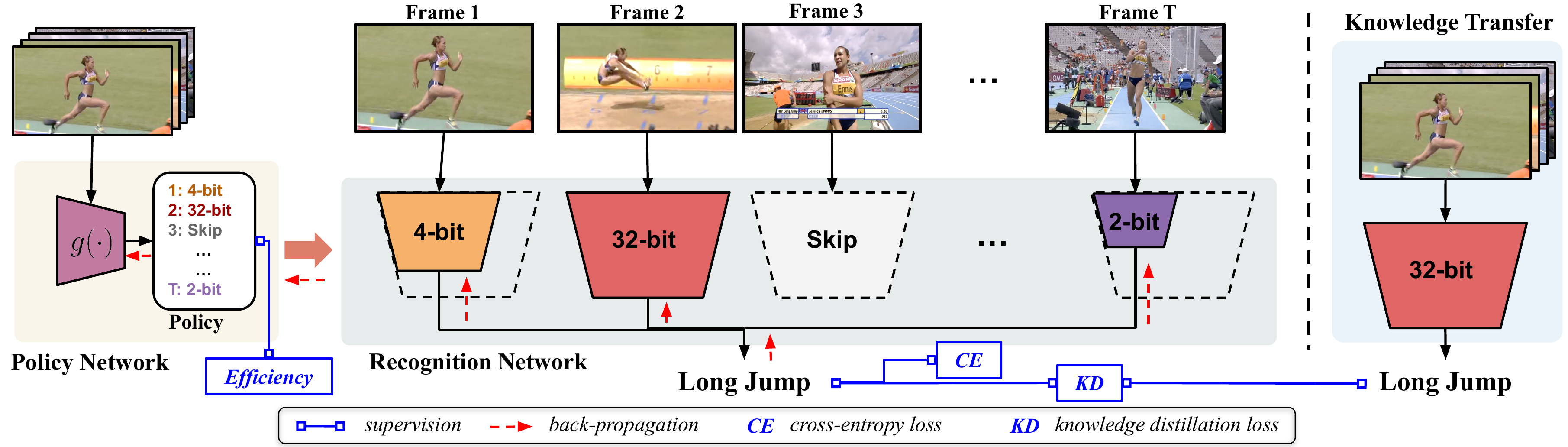}
\end{center} \vspace{-12pt}
   \caption{\small \textbf{Illustration of our proposed approach}. \ours consists of a very lightweight policy network and a single backbone network for recognition which can be simply quantized to lower precisions by truncating the least significant bits.
   The policy network decides what quantization precision to use on a per frame basis, in pursuit of a reduced overall computational cost without sacrificing recognition accuracy. We train both networks using back-propagation with a combined loss of standard cross-entropy and efficiency for video recognition. We additionally distill knowledge from a pre-trained full-precision model to guide the training of lower precisions. During inference, each frame is sequentially fed into the policy network to select optimal precision for processing the current frame through the recognition network and then the network averages all the frame-level predictions to obtain the video-level prediction.
   Best viewed in color.
   }
   \vspace{-7pt}
   \label{fig:our_model}
\end{figure*}

\vspace{1mm}
\noindent\textbf{Dynamic Computation.} Dynamic computation %
to improve efficiency has been studied from multiple perspectives~\cite{bengio2015conditional,bengio2013estimating,veit2018convolutional,wang2018skipnet,guo2019spottune,najibi2019autofocus,figurnov2017spatially,mcgill2017deciding}. Representative methods for image classification, dynamically adjust network depth~\cite{figurnov2017spatially,mcgill2017deciding,wu2018blockdrop,huang2017multi,yang2020mutualnet}, width~\cite{yu2018slimmable,chen2019you,hua2019channel}, perform routing~\cite{jie2019anytime,mcgill2017deciding} or switch resolutions~\cite{yang2020resolution}. 
Similar in spirit, dynamic methods for efficient video recognition adaptively select salient frames/clips~\cite{yeung2016end,wu2019adaframe,korbar2019scsampler,fan2018watching,wu2019multi,hussein2020timegate}, utilize audio~\cite{gao2020listentolook}, reduce feature redundancy~\cite{pan2021vared2}, or select frame resolutions~\cite{wu2019liteeval,meng2020ar}. Recently, AdaFuse~\cite{meng2021adafuse} proposes adaptive fusion of channels from current and past feature maps on a per instance basis, for recognizing video actions. 
Our approach is closely related yet orthogonal to these approaches as it focuses on network quantization to dynamically select the optimal bit-width conditioned on inputs, in pursuit of computational efficiency without sacrificing accuracy. Moreover, unlike existing works, our framework requires neither complex RL policy gradients~\cite{wu2019adaframe,wu2019multi,yeung2016end} nor additional modalities such as audio~\cite{gao2020listentolook,korbar2019scsampler} to learn dynamic policies. 

\vspace{1mm}
\noindent\textbf{Network Quantization.} Low-precision networks~\cite{han2015deep,zhou2016dorefa,choi2018pact}, have attracted intense attention in recent years. Early works such as~\cite{han2015deep,leng2018extremely,zhou2016dorefa} mainly focus on quantizing weights while using 32-bit activations. Recent approaches quantize both weights and activations through using uniform quantization that uses identical bit-width for all layers~\cite{zhang2018lq,choi2018pact,park2017weighted}, or mixed precision quantization that uses different bit-widths for different layers or even channels~\cite{wang2019haq,cai2020rethinking,wu2018mixed}. Binary networks~\cite{hubara2016binarized,rastegari2016xnor} constrain both weights and activations to binary values, which brings great benefits to specialized hardware devices. Designing efficient strategies for training low-precision~\cite{zhuang2018towards,kim2019qkd,zhuang2020training} or any-precision networks~\cite{jin2020adabits,yu2019any} that can flexibly adjust the precision during inference is also another recent trend in quantization. Despite recent progress, the problem of quantization for video recognition models is rarely explored. Moreover, existing methods perform quantization in a static manner with a fixed computational cost, leaving adaptive quantization conditioned on inputs an open problem.

\section{Proposed Method}

Given $T$ sampled frames from a video $V = \{x_1, x_2, \cdots, x_T\}$ with the action label $y$ and a set of $n$ candidate bit-widths (precisions) $\mathcal{B}=\{b_1,b_2, \cdots, b_n\}$ (assuming $b_1> b_2>\dots> b_n$), our goal is to seek (1) a policy function $g: V \rightarrow \mathcal{B}^T$ that automatically decides the optimal bit-width $b$ for the frame $x_i$ for processing in the recognition network, (2) a single recognition network $f: V \rightarrow y$ which can be quantized to different precisions in $\mathcal{B}$ without additional storage or computation cost.  With the desired policy network $g$ and recognition network $f$, our main objective is to improve accuracy,  while  taking the resource efficiency into account for video action recognition.
Note that given the optimal bit-width $b$ for the frame $x_i$, we quantize all the network weights and activations to the same bit-width $b$, which is well supported by existing hardwares.

\subsection{Preliminaries}\label{sec:Preliminaries}
We denote the full-precision network weights by $\bf{W}$ and activations by $\mathbf{A}$. Given a certain precision with bit-width $b$ and a quantization function $Q$, we denote the quantization of $\mathbf{W}$ and $\mathbf{A}$ as $Q(\mathbf{W}, b) = \widehat{W}_b$ and  $Q(\mathbf{A}, b) = \widehat{A}_b$. In this paper, we use DoReFa~\cite{zhou2016dorefa} for weight quantization and PACT~\cite{choi2018pact} for activation quantization.

\vspace{1mm}
\noindent\textbf{Weight Quantization.} 
DoReFa~\cite{zhou2016dorefa} normalizes $\mathbf{W}$ into $[-1, 1]$ and then rounds it to the nearest quantization levels:
\begin{align}
    & \widehat{W}_b = 2 \times \text{quantize}_b(\frac{tanh(\mathbf{W})}{2 \max tanh(\mathbf{W})} + \frac{1}{2}) -1, \label{eq:old_quantize1} \\
    & \text{quantize}_b(x) = \frac{1}{2^b -1} \times \lfloor (2^b - 1) x\rceil, \label{eq:old_quantize}
\end{align}
where $\lfloor . \rceil$ is the rounding operation. 

\vspace{1mm}
\noindent\textbf{Activation Quantization.} 
PACT~\cite{choi2018pact} introduces a learnable clipping value $\alpha$ for activations in each layer. More specifically, the activation $\mathbf{A}$ is first clipped into $[0, \alpha]$ and then rounded to the nearest quantization levels:
\begin{align}
    & \widehat{A}_b = \alpha \times  \text{quantize}_b (\text{clip}(A, 0, \alpha) /\alpha). \nonumber
\end{align}

\subsection{Approach Overview}\label{sec:approachOverview}

Figure~\ref{fig:our_model} shows an overview of our approach. In general, we learn a instance-specific policy $a_i$ that decides on-the-fly which precision to use (or even skip) for processing the current frame $x_i$, and a video classifier $f$ which can be flexibly quantized to the desired precision of the current frame by simply truncating the least significant bits without any extra computation or memory cost.  
To this end, \ours consists of a lightweight policy network $g$ and a video recognition network $f$. The policy network $g$ contains a feature extractor and an LSTM module 
to learn the discrete decisions of which precision to use, per input frame (see Section~\ref{sec:learningWithAdaptiveQuantization}).
Moreover, it is often unnecessary and inefficient to process every frame in a video due to large redundancy resulting from static scenes or frame quality being very low. Thus, we skip frames (i.e., precision set to zero) in addition to dynamic selection of precisions in an unified framework to improve efficiency in video recognition. To further enable flexible and scalable quantization, we learn the video classifier as an any-precision network and design a simple yet effective optimization scheme to ensure that the single set of network weights get executed with multiple precisions without additional storage and computation cost (see Section~\ref{sec:learningAdjustablePrecision}).

During the training, we first learn the any-precision recognition network and then optimize the policy network with Gumbel-Softmax Sampling~\cite{jang2016categorical} through standard back-propagation. 
We design the loss to achieve both competitive performance and computational efficiency (measured by FLOPS~\cite{wang2019learning}) required for video recognition.
We additionally distill knowledge from a pre-trained full-precision model to guide training of the lower precisions.
During the inference, each video frame is sequentially fed into the policy network whose output decides the right precision to use for the given frame and then the frame is processed through the recognition network with the predicted precision to generate a frame-level prediction. Finally, the network averages predictions of all the frames as the final video-level prediction. It is worth noting that the policy network is designed to be very lightweight so that its computational overhead is negligible (e.g., MobileNetv2~\cite{sandler2018mobilenetv2} in our work). 

\subsection{Learning Dynamic Quantization Policy}\label{sec:learningWithAdaptiveQuantization}
\ours learns the frame-wise policy $a_i$ to decide which precision to process the frame $x_i$ or directly skip it where skipping can be viewed as processing the frame with $0$-bit. So our entire action space is $\Omega = \mathcal{B} \cup \{0\}$. We generate decision $a_i \in \Omega, \forall i \in [1, T]$ from the policy network $g$ sequentially. We compose the policy network with a feature extractor $\phi$ followed by an LSTM module: 
\begin{align}
 h_i, o_i = \text{LSTM}(\phi(x_i), h_{i-1}, o_{i-1}),
\end{align}
where $h_i$ and $o_i$ are hidden state and outputs of LSTM at the time step $i$.
We further compute the distribution $\pi_i \in \mathbb{R}^{|\Omega |}  $ over our action space $\Omega$ from $h_i$:
\begin{align}
    \pi_i = \text{Softmax} (\textit{fc} (h_i)).
\end{align}

However, sampling policy $a_i$ from the discrete distribution $\pi_i$ is non-differentiable which makes direct optimization difficult. One way to solve this is to model the optimization problem as a reinforcement learning problem and then derive the optimal parameters of the policy network using policy gradient methods~\cite{williams1992simple}. However, policy gradient is often complex, unwieldy to train and requires techniques to reduce variance during training as well as carefully selected reward functions. In contrast, we use Gumbel-Softmax Sampling~\cite{jang2016categorical} to circumvent this non-differentiability and make our framework fully differentiable, as in~\cite{wu2019liteeval, sun2020adashare}.

\vspace{1mm}
\noindent\textbf{Gumbel-Softmax Sampling.}  The Gumbel Softmax trick~\cite{jang2016categorical} substitutes the original non-differentiable sample from a discrete distribution with a differentiable sample from a corresponding Gumbel-Softmax distribution. 

Specifically, instead of directly sampling $a_i$ from its distribution $\pi_i$, we generate it as, 
\begin{align}
    a_{i} = \argmax_{j\in \Omega} \big(\log\pi_{i}(j) + G_{i}(j)\big),~\label{eq:argmax_gumbel}
\end{align}
where $G_{i}=-\log(-\log U_i)$ is a standard Gumbel distribution with $U_{i}$ sampled from a uniform distribution $\text{Unif}(0,1)$.
To remove the non-differentiable argmax operation in Eq.~\ref{eq:argmax_gumbel}, the Gumbel Softmax trick relaxes $\text{one-hot}(a_{i}) \in \{0, 1\}^{|\Omega|}$ (the one-hot encoding of $a_i$) to $p_{i} \in \mathbb{R}^{|\Omega|}$ with the reparameterization trick~\cite{jang2016categorical}:
\begin{align}
p_{i}(j) = \frac{\exp\big((\log\pi_{i}(j) + G_{i}(j)) / \tau\big)}{\sum\limits_{k \in \Omega }\exp\big((\log\pi_{i}(k) + G_{i}(k)) / \tau\big)}, ~\label{eq:softmax_gumbel}
\end{align}
where $j \in \Omega $ and $\tau$ is the temperature of the softmax. Clearly, when $\tau>0$, the Gumbel-Softmax distribution $p_{i}$ is smooth so $\pi_{i}$ can be directly optimized by gradient descent, and when $\tau$ approaches 0, the soft decision $p_i$ becomes the same as $\text{one-hot}(a_i)$. Following~\cite{guo2019spottune,sun2020adashare}, we set $\tau=5$ as the initial value and gradually anneal it down to 0 during training.  

\subsection{Any-Precision Video Recognition}\label{sec:learningAdjustablePrecision}
Given frame-specific precisions, quantizing weights and activations of a single network while recognizing videos is a major challenge. A naive strategy is to manually train different models tailored for the different precision and then route frames to the corresponding models to generate predictions. However such a strategy requires time-consuming training for each of the models and also increases the memory storage cost, making it inefficient for many real-time applications. To tackle this problem, we adopt any-precision recognition~\cite{jin2020adabits,yu2019any} that makes a single model be flexible to any numerical precision during the inference. Specifically, we first modify the weight quantizer to enable the network parameters to get quantized to lower precision with low computation cost after the training. Then, we propose a simple and effective learning scheme for training of the any-precision video recognition network.

With the original DoReFa quantization~\cite{zhou2016dorefa} (Eq.~\ref{eq:old_quantize1} and \ref{eq:old_quantize}), all numerical precisions need to be quantized down from the full-precision value. Thus, the repeated weight quantizations cause redundant computation when the recognition network frequently switches across different precisions. To reduce computational cost of switching operation, we quantize full precision weight $W$ to the largest bit-width $b_1$ and then truncate least significant $b_1 - b$ bits to get quantized weight $\widehat{W}_{b}$. We save the quantized $b_1$-bit network weights after the training. Benefiting from this modified quantization, we only need to discard the extra bits to switch to lower precisions during inference. Furthermore, we align $\mathbb{E}[\widehat{W}_b]$ with $\mathbb{E}[\widehat{W}_{b_1}]$ to minimize the mean discrepancy caused by discarded bits.

Inspired by~\cite{yu2018slimmable,jin2020adabits}, we jointly train a single network under different bit-widths with shared weights for any-precision video recognition. Specifically, we gather losses of all precisions with same input batch and then update the network. To get the loss of a precision with bit-width $b$, we feed the input video and quantize network weights and activations to $b$-bit for every frame. To resolve mismatch in statistics of activations with different precisions, we use a separate set of Batch Normalization layers and clipping level parameters for different precisions~\cite{yu2018slimmable}. Moreover, following the success of knowledge distillation~\cite{hinton2015distilling}, we transfer knowledge from a pretrained full-precision recognition network to guide training of lower precisions because the full-precision weights is expected to give confident predictions, and provide valuable knowledge in its soft logits, while the low-precision student gains the knowledge by mimicking the teacher.

\subsection{Losses}\label{sec:losses}

For video action recognition, we minimize standard cross-entropy loss between predicted label and ground truth action: 
\begin{align}
    \mathcal{L}_{ce} (V|A) = \mathbb{E}[-y \log (f(V | A))],
\end{align}
where $A = {a_1, a_2, \cdots, a_T}$ represents precisions to use for the sampled $T$ frames, which can be either predicted by the lightweight policy network ($A=g(V)$) or set manually. 

To better guide the optimization of the model with lower capacity, \eg the recognition network with lower precision, we utilize a distillation loss $\mathcal{L}_{kd}$ to transfer knowledge from a pretrained full-precision video recognition network (teacher) by taking Kullback–Leibler (KL) divergence between soft-logits of our model $y_{A}$ and of the teacher network $y_t$ as
\begin{align}
     \mathcal{L}_{kd} (V|A) =  \text{KL}(y_t || y_A) = \sum_{i = 1}^{m} (y_t)_i \log \frac{(y_t)_i }{(y_A)_i },  
\end{align}
where $m$ is the number of video categories and $(\cdot)_i$ denotes the $i$-th element of the vector.
Thus, given the input video $V$, the overall loss $\mathcal{L}_{f}$ to optimize the any-precision video recognition network $f$ is defined as
\begin{align} \label{eq:any}
    \mathcal{L}_{f} (V) = \sum_{A = b_1 ^T , \cdots b_n^T}  \mathcal{L}_{ce} (V|A)  + \mathcal{L}_{kd} (V|A). 
\end{align}

To address computational efficiency, we pre-compute FLOPs~\cite{wang2019learning} needed for one frame to get processed in the recognition network with different candidate precisions in $\mathcal{B}$. We directly minimize FLOPs usage per video with the generated policy $A$, to reduce the computational cost as
\begin{align}
    \mathcal{L}_{e} (A) = \sum_{i=1}^ {T} (\text{FLOP}(a_i)).
\end{align}

Furthermore, we introduce two additional regularizers to better optimize the policy network. First, we enforce a balanced policy usage over the entire action space to avoid the policy network learning some sub-optimal solutions where some actions are totally ignored. More formally, we define the balanced policy usage loss $\mathcal{L}_{b}$ as  
\begin{align}
    \mathcal{L}_{b} (A) = \sum_{k \in \Omega} (\mathbb{E} [\frac{1}{T} \sum_{i=1} ^ {T}   \mathbbm{1} (a_i = k) ] - \frac{1}{|\Omega|}  ).
\end{align}

Second, we minimize the entropy of the learned probability distribution over the action space $\Omega$ of each frame. It forces the policy network to avoid randomness during the inference by generating deterministic prediction for the precision to use for each video frame:
\begin{align}
    \mathcal{L}_{d} (\pi) = \sum_{i=1} ^ {T} H(\pi_i), 
\end{align}
where $H(\cdot)$ is the entropy function.
Finally, the overall loss $\mathcal{L}_{g}$ to optimize the policy network $g$ is defined as
\begin{align} \label{eq:policy}
    \mathcal{L}_{g} (V)  = \ & \mathcal{L}_{ce} (V|A)  + \mathcal{L}_{kd} (V|A) \nonumber \\
    & + w_1  \mathcal{L}_{e} (A) + w_2  \mathcal{L}_{b} (A) + w_3  \mathcal{L}_{d} (\pi),
\end{align}
where $A=g(V)$, and $w_1$, $w_2$ and $w_3$ are hyperparameters to balance loss terms. 
In summary, we first jointly train the any-precision recognition network $f$ with all precisions in $\mathcal{B}$ (using Eq.~\ref{eq:any}), and then train policy network $g$ (using Eq.~\ref{eq:policy}) to generate policy over the action space $\Omega$ per input frame.

\section{Experiments} \label{sec:experiments}

\subsection{Experimental Setup}

\noindent\textbf{Datasets.} We evaluate our approach using four datasets, namely ActivityNet-v1.3~\cite{caba2015activitynet}, FCVID~\cite{jiang2017exploiting}, Mini-Sports1M~\cite{karpathy2014large} and Mini-Kinetics~\cite{carreira2017quo}. 
ActivityNet contains $10,024$ videos for training and $4,926$ videos for validation across $200$ categories. FCVID consists of $45,611$ videos for training and $45,612$ videos for testing across $239$ classes. Mini-Sports1M~\cite{gao2020listentolook} is a subset of full Sports1M dataset~\cite{karpathy2014large} containing $30$ videos per class in training and $10$ videos per class in testing over $487$ classes. Mini-Kinetics~\cite{chen2020deep} is a subset of full Kinetics400~\cite{carreira2017quo} dataset containing $121,215$ videos for training and $9,867$ videos for testing across $200$ classes. 

\vspace{1mm}
\noindent\textbf{Implementation Details.}
We adopt temporal segment network (TSN)~\cite{wang2016temporal} to aggregate the predictions over $T=16$ uniformly sampled frames from each video. 
We use ResNet-18 and ResNet-50~\cite{he2016deep} for the recognition network while MobileNetv2 \cite{sandler2018mobilenetv2} combined with a single-layer LSTM (with 512 hidden units) to serve as policy network in all our experiments. To save computation, we use lower resolution images ($84 \times 84$) in  policy network. We set the action space $\Omega = \{32, 4, 2, 0\}$ in all experiments, i.e., the policy network can choose either one out of $\{32, 4, 2\}$ precision or skip frame for efficient recognition.
We first train the any-precision recognition network (pretrained from ImageNet weights) for 100 epochs to provide a good starting point for policy learning and then train the policy network for 50 epochs on all datasets. 
We use separate sets of learning parameters (learning rate, weight decay) for clipping values of each precision.
Following ~\cite{zhou2016dorefa, choi2018pact}, we do not quantize input, first layer and last layer of the network. 
More implementation details are included in the Appendix~\ref{sec:impl}.

\vspace{1mm}
\noindent\textbf{Baselines.} We compare our approach with the following baselines and existing approaches. First, we consider a 2D-CNN based \enquote{Uniform} baseline that uses 32-bit precision to process all the sampled frames and then averages the frame-level results as the video-level prediction. We also compare with two more variants of uniform baseline that uses lower precisions such as 4-bit and 2-bit respectively to process the video frames.
Second, we compare with \enquote{Ensemble} baseline that gathers all the frame-level predictions by processing them at different precision (instead of selecting an optimal precision per frame). This serves as a very strong baseline for classification, at the cost of heavy computation. Finally, we compare our method with existing efficient video recognition approaches, including LiteEval~\cite{wu2019liteeval} (NeurIPS'19), SCSampler~\cite{korbar2019scsampler} (ICCV'19), AR-Net~\cite{meng2020ar} (ECCV'20), and AdaFuse~\cite{meng2021adafuse} (ICLR'21). 
We directly quote the numbers reported in the published papers when possible or use authors provided source codes~\cite{wu2019liteeval,meng2021adafuse} using the same backbone and experimental settings for a fair comparison. 

\vspace{1mm}
\noindent\textbf{Metrics.} 
We  compute  either mAP (mean average precision) or Top-1 accuracy depending on datasets to measure performance of different methods. We follow~\cite{wang2019learning,phan2020binarizing,shen2020once} and measure computational cost with giga floating-point operations (GFLOPs), which is a hardware independent metric. 
Specifically, given FLOPs of a full-precision layer by $\mathbf{a}$, the FLOPs of $m$-bit weight and $n$-bit activation quantized layer is  $\frac{mn}{64} \times \mathbf{a}$.
We also measure memory usage (MB) represented by the storage for parameters of the network, as in~\cite{wang2019learning}.

\begin{table}
    \begin{center}
        \resizebox{1\linewidth}{!}{
        \begin{tabular}{c|cc|cc|c}
            \Xhline{3\arrayrulewidth} 
            \multirow{2}{*}{Model} &  \multicolumn{2}{c|}{ActivityNet} & \multicolumn{2}{c|}{FCVID}    & \multirow{2}{*}{ \makecell{Mem. \\ (MB)}}  \\
             \cline{2-5} 
              &  mAP (\%) & GFLOPs  &  mAP (\%) & GFLOPs &  \\
             \Xhline{3\arrayrulewidth} 
             \multicolumn{6}{c}{ResNet-18} \\
             \Xhline{3\arrayrulewidth}
            Uniform (32-bit) & 69.7 & 29.1  & 77.6 & 29.1 & 43.1 \\
            Uniform (4-bit) & 68.0 &  7.3 &  76.5 & 7.3 & 5.4 \\
            Uniform (2-bit) & 65.2 & 1.8  & 74.3 & 1.8 & 2.7 \\
            Ensemble & 70.7 & 38.2  & 78.8 & 38.2 & 51.2 \\
            \hline 
            \ours & 70.9 & 9.5  & 79.1 & 9.4  & 50.2 \\
             \Xhline{3\arrayrulewidth} 
             \multicolumn{6}{c}{ResNet-50} \\
             \Xhline{3\arrayrulewidth} 
            Uniform (32-bit) & 72.5 & 65.8  & 81.0 & 65.8 & 91.4 \\
            Uniform (4-bit) & 71.7 & 16.5  & 79.3 & 16.5 & 11.4 \\
            Uniform (2-bit) & 69.3 & 4.1  & 78.5 & 4.1 & 5.7 \\
            Ensemble & 74.7 & 86.4 & 83.0 & 86.4 & 108.5\\
            \hline 
            \ours & 74.8 & 28.1  & 82.7 & 27.0 & 98.6 \\
            \Xhline{3\arrayrulewidth} 
        \end{tabular}
        } \vspace{-2mm}
         \caption{\small \textbf{Video recognition results on ActivityNet and FCVID}. Our approach \ours outperforms all the simple baselines.}
        \label{table:simple_baselines_actv_fcvid}
    \end{center} \vspace{-7mm}
\end{table}

\subsection{Results and Analysis}

\noindent\textbf{Comparison with Traditional Uniform Baselines.} We first compare \ours using different backbones (ResNet-18 and ResNet-50) to show how much performance our dynamic approach \ours can achieve compared to simple 2D-CNN based baselines on both ActivityNet and FCVID datasets.
As shown in Table~\ref{table:simple_baselines_actv_fcvid}, our approach consistently outperforms the full-precision uniform baseline (32-bit) in both mAP and GFLOPS, with minimal increase in memory on both datasets. Using ResNet-18 as the backbone, \ours obtains an mAP of $70.9\%$ and $79.1\%$, requiring $9.5$ and $9.4$ GFLOPS on ActivityNet and FCVID respectively. Uniform quantization with low bit-widths leads to a significant reduction in computation and memory but they suffer from a noticeable degradation in recognition performance, e.g., the 2-bit performance is $4.5\%$ and $3.3\%$ lower than the 32-bit counterpart on ActivityNet and FCVID respectively. 

Similarly, with ResNet-50, \ours offers $56.7\%$ ($65.8$ vs $28.1$) and $58.9\%$ ($65.8$ vs $27.0$) savings in GFLOPS while outperforming the Uniform (32-bit) baseline by $2.1\%$ and $2.7\%$ in mAP on ActivityNet and FCVID, respectively. We further compare with 8-bit Uniform Baseline that uses same percentage of random skipping as \ours (i.e. $8\%$ random skipping on ActivityNet). 
With ResNet-50, our approach outperforms this baseline by $2.7\%$ ($72.1\%$ vs $74.8\%$), showing effectiveness of learned policy in selecting optimal quantization precision per frame while recognizing videos. 

As shown in Table~\ref{table:simple_baselines_actv_fcvid}, Ensemble achieves comparable recognition performance because it is a very strong baseline that gathers all the predictions by processing frames through multiple backbones. However, \ours provides $67.4\%$ and $68.7\%$ computational savings including a $10\%$ savings in memory over the Ensemble baseline on ActivityNet and FCVID respectively, showing the importance of instance-aware dynamic quantization for efficient video recognition. Moreover, we also compare with a Weighted Ensemble baseline, where weights are assigned based on entropy of softmax scores to reflect prediction confidence of different predictions. We observe that it only achieves $0.3\%$ higher mAP while requiring $67.4\%$ more computation than our method on ActivityNet ($75.1\%$ vs $74.8\%$). Note that \ours requires less computation on average on FCVID than ActivityNet as FCVID contains more static videos with high redundancy compared to ActivityNet that consists of action-centric videos with rich temporal information. 
\begin{table}
    \begin{center}
     
        \resizebox{1\linewidth}{!}{
        \begin{tabular}{c|cc|cc|c}
             \Xhline{3\arrayrulewidth} 
             \multirow{2}{*}{Model} &  \multicolumn{2}{c|}{ActivityNet} & \multicolumn{2}{c|}{FCVID}    & \multirow{2}{*}{\makecell{Mem. \\ (MB)}}  \\
             \cline{2-5} 
              &  mAP (\%) & GFLOPs  &  mAP (\%) & GFLOPs &  \\
             \Xhline{3\arrayrulewidth}  
            LiteEval & 72.7 & 95.1 & 80.0 & 94.3 & 177.2  \\
            SCSampler & 72.9 & 42.0 & 81.0 & 42.0 & 98.6 \\
            AR-Net & 73.8 & 33.5 & 81.3 & 35.1 & 223.4 \\
            AdaFuse &  73.1 & 61.4 & 81.6 & 45.0 & 151.2 \\ %
            \hline 
            \ours & \textbf{74.8} & \textbf{28.1} & \textbf{82.7} & \textbf{27.0} & \textbf{98.6} \\
            \Xhline{3\arrayrulewidth} 
        \end{tabular}
        } \vspace{-2mm}
        \caption{\small \textbf{Comparison with state-of-the-art methods on ActivityNet and FCVID}. \ours achieves the best mAP while offering significant savings in both GFLOPS and Memory (MB).}
        \label{table:sota_actv_fcvid}
    \end{center} \vspace{-5mm}
\end{table}

\begin{table}
    \begin{center}
    
        \resizebox{1\linewidth}{!}{
       \begin{tabular}{c|cc|cc|c}
             \Xhline{3\arrayrulewidth} 
             \multirow{2}{*}{Model} &  \multicolumn{2}{c|}{Mini-Sports1M} & \multicolumn{2}{c|}{Mini-Kinetics}    & \multirow{2}{*}{\makecell{Mem. \\ (MB)}}  \\
             \cline{2-5} 
              &  mAP (\%) & GFLOPs  &  Tops-1 (\%) & GFLOPs &  \\
             \Xhline{3\arrayrulewidth}  
            LiteEval & 44.7 & 66.2 & 61.0 & 99.0 & 177.2 \\
            SCSampler & 44.3 & 42.0  & 70.8 & 42.0 & 98.6 \\
            AR-Net & 45.0 &	37.6 & 71.7 & 32.0 & 223.4 \\
            AdaFuse & 44.1 & 60.3 & 72.3 & 23.0 & 151.2 \\ %
            \hline 
            \ours & \textbf{46.4} & \textbf{26.8} & \textbf{72.3} & \textbf{20.4} & \textbf{98.6}\\
            \Xhline{3\arrayrulewidth} 
        \end{tabular}
        } \vspace{-2mm}
         \caption{\textbf{Comparison with state-of-the-art methods on Mini-Sports1M and Mini-Kinetics}. Our approach \ours (w/ ResNet-50) obtains the best performance with great savings in computation (GFLOPS) and memory (MB).}
        \label{table:sota_sports1m_kinetics}
    \end{center} \vspace{-7mm}
\end{table}

\begin{figure}
\begin{center}
     \includegraphics[width=0.85\linewidth]{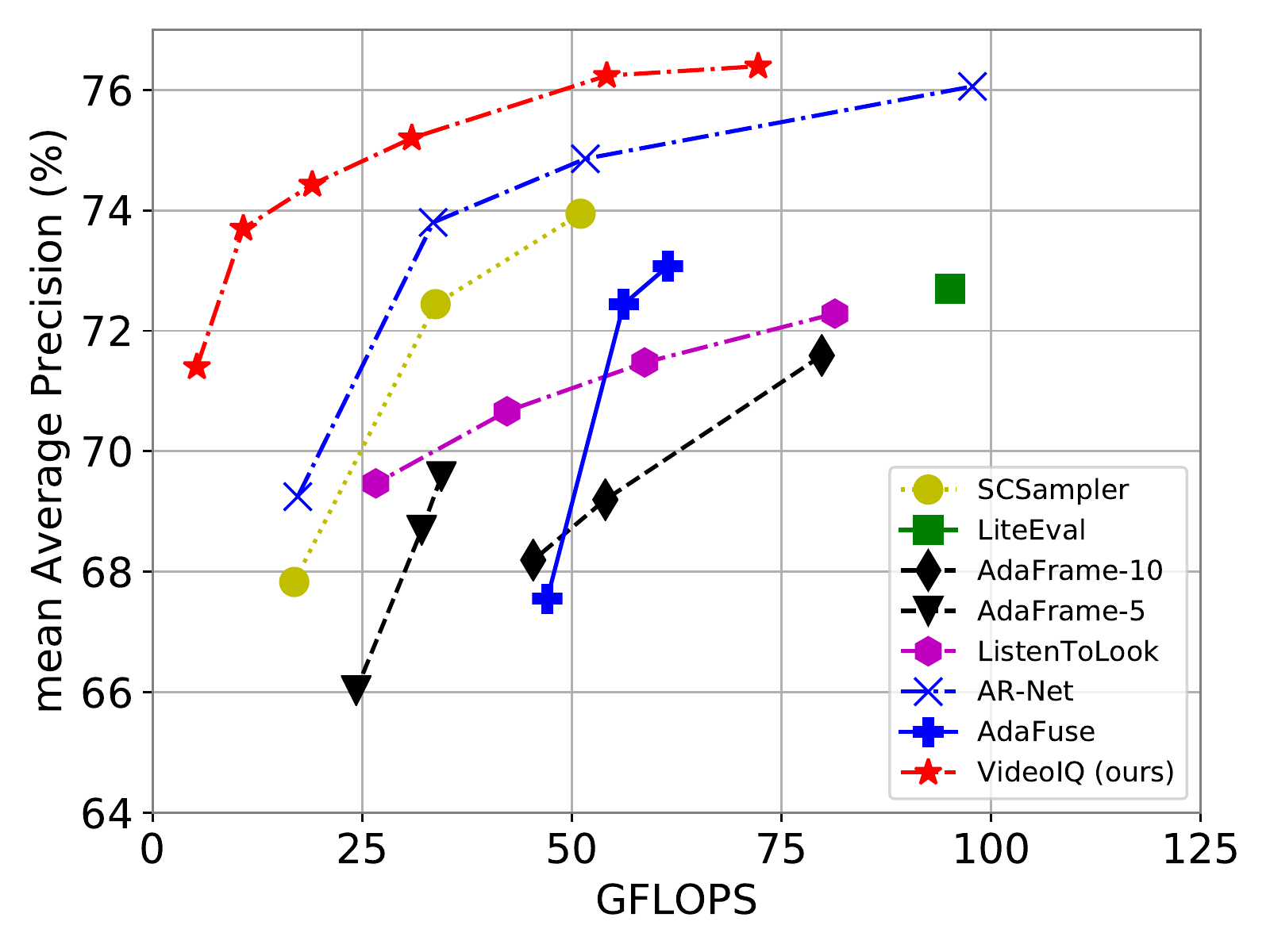}
\end{center} \vspace{-6mm}
   \caption{\small \textbf{Computational cost (GFLOPS) vs mean Average Precision (\%) on ActivityNet dataset}. 
   \ours (red points) achieves the best trade-off when compared to existing methods.
   }
   \vspace{-3mm}
   \label{fig:map_vs_gflops}
\end{figure}

\vspace{1mm}
\noindent\textbf{Comparison with State-of-the-Art Methods.} 
Tables~\ref{table:sota_actv_fcvid}-\ref{table:sota_sports1m_kinetics} summarize the results and comparisons with existing dynamic inference methods on all four datasets. 
Our approach is clearly better than all the compared methods in terms of both accuracy and resource efficiency (computation and memory), making it suitable for efficient video recognition. \ours obtains an mAP (accuracy for Mini-Kinetics) of $74.8\%$, $82.7\%$, $46.4\%$ and $72.3\%$, while requiring $28.1$, $27.0$, $26.8$ and $20.4$ GFLOPs on ActivityNet, FCVID, Mini-Sports1M and Mini-Kinetics, respectively. 
Note that while most of the compared methods reduce computation at the cost of significant increase in memory, our approach improves computational efficiency by using a model whose memory size is just slightly larger than the 32-bit model. 

Among the compared methods, AR-Net is the most competitive in terms of computational efficiency. However, \ours consistently outperforms AR-Net in recognition performance while providing $26.0\%$ savings on average in computation and $55.8\%$ savings in memory. This is because of our two introduced components working in concert: dynamic quantization for computational efficiency and use of a single any-precision recognition network instead of separate models for memory efficiency. Likewise when compared with the recent method AdaFuse, our approach offers an average $41.1\%$ and $34.7\% $ reduction in computation and storage memory while improving the recognition performance (maximum $2.3\%$ on Mini-Sports1M) across all the datasets. AdaFuse obtains the best performance compared to other existing methods on Mini-Kinetics but it fails to achieve similar performance on untrimmed video datasets. We suspect that being a method that relies on efficient reuse of history feature maps, it fails to aggregate the information of all time stamps when the video gets very long, as in untrimmed datasets. In summary, \ours establishes new state-of-the-art for the task of efficient video recognition on four datasets, improving previous best result in terms of accuracy, computational efficiency and memory efficiency.  

Figure~\ref{fig:map_vs_gflops} compares our approach to the existing methods by varying computational budgets on ActivityNet. Our method consistently outperforms all the compared methods and achieves the best trade-off between computational cost and accuracy, which once again shows that \ours is an effective and efficient design for video recognition.

\begin{table}
    \begin{center}
        \resizebox{1\linewidth}{!}{
       \begin{tabular}{c|cccc}
             \Xhline{3\arrayrulewidth} 
         \backslashbox{Train}{Test} & ActivityNet & FCVID  & Mini-Sports1M & Mini-Kinetics \\
             \Xhline{3\arrayrulewidth}  
            ActivityNet & \textcolor{blue}{\textbf{74.8}} &  82.7 &  46.3 &  71.6 \\
            FCVID & 74.4 &  \textcolor{blue}{\textbf{82.8}} &  45.8 &  72.1\\
            Mini-Sports1M & 74.6 &  82.6 &  \textcolor{blue}{\textbf{46.4}} & 72.2 \\
            Mini-Kinetics & 74.7 & 82.7 &  46.3 &  \textcolor{blue}{\textbf{72.3}} \\
            \Xhline{3\arrayrulewidth} 
        \end{tabular}
        } \vspace{-2mm}
         \caption{\small \textbf{Transferring learned policies}. Diagonal numbers refer to training and testing the quantization policy on the same dataset while non-diagonal numbers refer to learning the policy on one dataset (rows) and testing on others (columns).}
        \label{table:zero_shot}
    \end{center} \vspace{-7mm}
\end{table}

\vspace{1mm}
\noindent\textbf{Transferring Learned Policies.} We analyze transferability of our learned policy by performing cross-dataset experiments, i.e., learning policy on one dataset while testing on the other. Specifically, we take the policy network trained on one dataset and utilize it directly for testing along with a trained any-precision recognition network on another dataset. Table~\ref{table:zero_shot} summarizes the results. As expected, training and testing on the same dataset provides the best performance on all cases (marked in blue). However, the negligible difference among the values across each column clearly shows that policies learned using our method are transferable to unseen classes and videos across different datasets.

\begin{figure*}
\begin{center}
     \includegraphics[width=\linewidth]{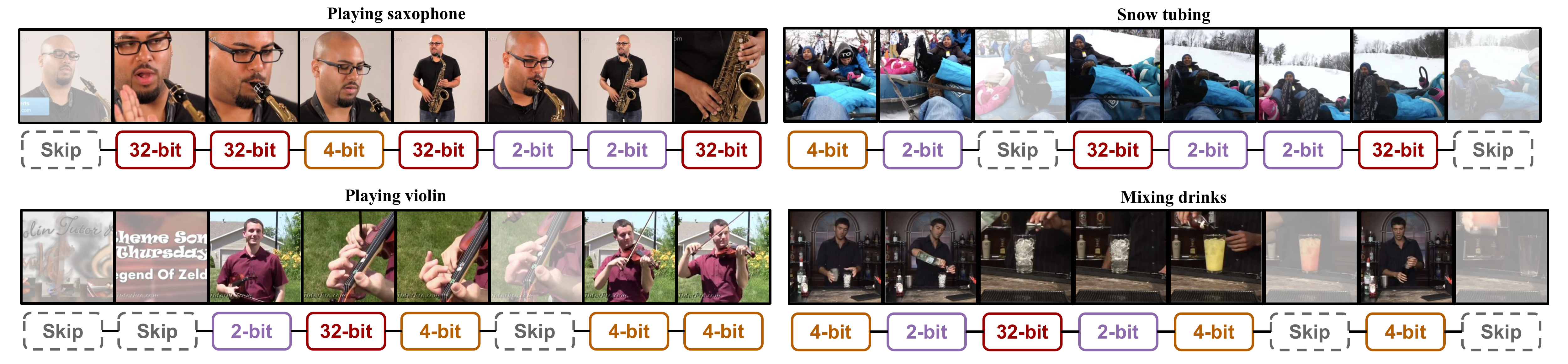}
\end{center} \vspace{-6mm}
   \caption{\small \textbf{Qualitative examples from ActivityNet dataset.} Our approach \ours processes more informative frames with high precision and less informative ones with lower precision or skip them when irrelevant, for efficient video recognition. Best viewed in color.  
   }
   \vspace{-3mm}
   \label{fig:qualitative results}
\end{figure*}

\begin{figure}
\begin{center}
     \includegraphics[width=\linewidth]{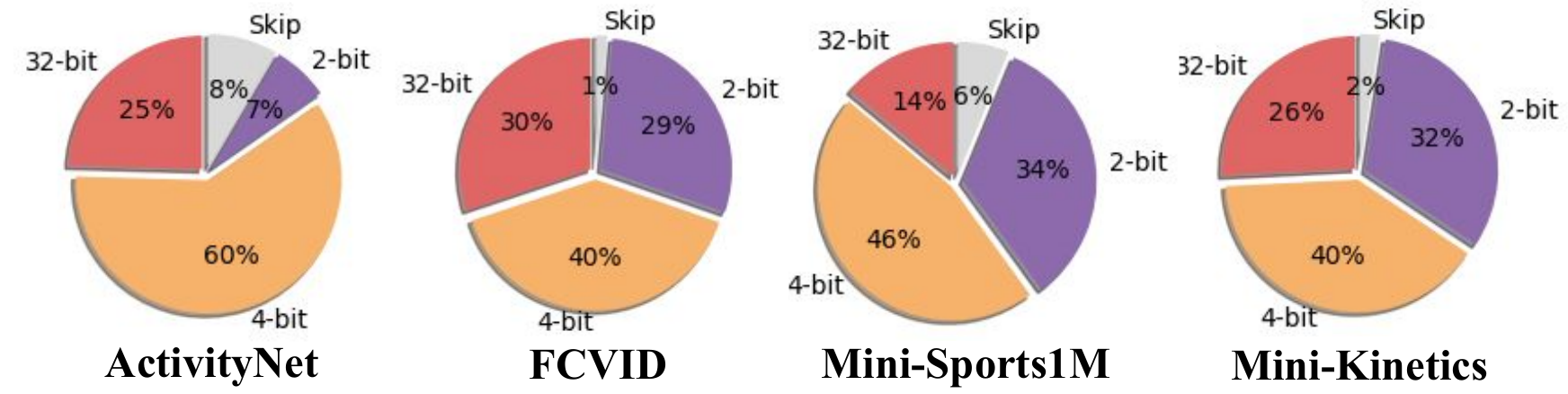}
\end{center} \vspace{-4mm}
   \caption{\small \textbf{Dataset-specific policy distribution}.
   }
   \vspace{-2mm}
   \label{fig:distribution_pie_chart}
\end{figure}

\vspace{1mm}
\noindent\textbf{Qualitative Analysis.} 
To better understand the learned policy, we visualize selected precision per input frame in Figure~\ref{fig:qualitative results}. Videos are uniformly sampled in 8 frames. Overall, our approach \ours focuses on the right quantization precision to use per frame for correctly classifying videos while taking efficiency into account. \ours processes the most indicative frames in 32-bit precision while it uses lower precision (or skips) for frames that irrelevant to the action (e.g., \enquote{Playing saxophone} and \enquote{Snow Tubing}). Similarly in the case of \enquote{Playing violin} and \enquote{Mixing drinks}, after being confident about the prediction, it interestingly avoids using the 32-bit precision even if informative content appear later in the video. More qualitative examples are included in the Appendix~\ref{sec:qual}.

Figure~\ref{fig:distribution_pie_chart} shows the overall policy distribution on different datasets. Our approach leads to distinctive policy patterns representing different characteristics of datasets. For example, while only few frames on ActivityNet use 2-bit precision, about $30\%$ of the frames on the other datasets can be processed using 2-bit precision, leading to different amount of computational savings across datasets. \ours skips very few frames on Mini-Kinetics ($2\%$), which is because Mini-Kinetics dataset contains short trimmed videos ($6-10$ seconds) while the remaining datasets consists of long untrimmed videos, lasting up to $5$ minutes.  

\subsection{Ablation Studies}

We present the following ablation experiments using ResNet-50 on ActivityNet dataset to show the effectiveness of different components in our proposed method.

\vspace{0.5mm}    
\noindent\textbf{Effect of Different Losses.} Table~\ref{table:ablation_loss} summarizes the effect of different losses on ActivityNet. Training without knowledge transfer from the 32-bit model (top row: by turning off $\mathcal{L}_{kd}$) only obtains a mAP of $73.5\%$ with similar GFLOPS as ours, which shows that it is important to utilize soft targets of the full-precision model as the teacher to guide lower precisions in learning. As expected, training by setting $\mathcal{L}_{e}$ to $0$ achieves the highest mAP of $75.1\%$ while requiring $38.5\%$ more GFLOPS compared to the one that uses efficient loss in training ($2^{nd}$ vs $3^{rd}$ row). Finally, adding both regularizations ($\mathcal{L}_{b}$ and $\mathcal{L}_{d}$) during the policy learning leads to the best performance with least computation showing the effectiveness of different losses in our framework.

\vspace{0.5mm}
\noindent\textbf{Effect of Decision Space. } We investigate the effect of decision space $\Omega$ by using different combinations of precision and skipping. 
As shown in Table~\ref{table:different_b}, only skipping frames (i.e., $\Omega = \{32, 0\}$) leads to an mAP of $72.9\%$ while setting the decision space to choose only precisions (i.e., $\Omega = \{32, 4, 2\}$) leads to an mAP of $74.5\%$ on ActivityNet. Compared to all the alternatives, the best strategy is to combine the set of precisions with skipping by setting $\Omega = \{32, 4, 2, 0\}$ for achieving top performance of $74.8\%$ in mAP with $28.1$ GFLOPS on ActivityNet dataset.

\begin{table}
    \begin{center}
        \resizebox{0.8\linewidth}{!}{
       \begin{tabular}{ccccc|cc}
             \Xhline{3\arrayrulewidth} 
              $\mathcal{L}_{ce}$ & $\mathcal{L}_{kd}$ & $\mathcal{L}_{e}$ & $\mathcal{L}_{b}$ & $\mathcal{L}_{d}$ &  mAP (\%) & GFLOPs   \\
             \Xhline{3\arrayrulewidth}  
           \checkmark & &  \checkmark  &  \checkmark  &   \checkmark & 73.5 & 29.0  \\
           \checkmark & \checkmark & & & & \textbf{75.1} & 56.4 \\
        \checkmark & \checkmark &\checkmark & & & 74.5 & 34.6 \\
            \checkmark & \checkmark &\checkmark &\checkmark & & 74.3 & 32.0 \\
            \checkmark & \checkmark &\checkmark &\checkmark & \checkmark & 74.8 & \textbf{28.1}\\
            \Xhline{3\arrayrulewidth} 
        \end{tabular}
        } \vspace{-2mm}
          \caption{\small \textbf{Effect of different losses on ActivityNet}.}
        \label{table:ablation_loss}
    \end{center} \vspace{-7mm}
\end{table}
\begin{table}
    \begin{center}
        \resizebox{0.8\linewidth}{!}{
       \begin{tabular}{c|cc}
             \Xhline{3\arrayrulewidth} 
              Decision Space $\Omega$ &  mAP (\%) & GFLOPs    \\
             \Xhline{3\arrayrulewidth}  
            \{32, 0\} &  72.9 & 31.6 \\
             \{32, 4, 2\} &  74.5 & 31.4 \\
            \{32, 4, 0\} & 74.7 & 32.8 \\
             \{32, 2, 0\} &  74.0 & 31.2 \\
            \{32, 4, 2, 0\} & \textbf{74.8} & \textbf{28.1}  \\
            \Xhline{3\arrayrulewidth} 
        \end{tabular}
        } \vspace{-2mm}
    \caption{\small \textbf{Effect of different decision space on ActivityNet}. Note that $0$ indicates skipping the frame for processing by the classifier.}
    \label{table:different_b}
    \end{center} \vspace{-7mm}
\end{table}

\vspace{0.5mm}
\noindent\textbf{Comparison with Random Policy.} We compare with random policy that uses the same backbone framework but randomly samples policy actions from uniform distribution and observe that our approach outperforms it by $2\%$ in mAP ($72.8\%$ vs $74.8\%$) on ActivityNet, which demonstrates effectiveness of learned policy in selecting optimal quantization precision per frame while recognizing videos. We also observe similar improvements ($\sim 2\%-3\%$) on other datasets.

\vspace{0.5mm}
\noindent\textbf{Effectiveness of Any-Precision Recognition Network.} 
We use three separate precision specific quantized models as part of the classifier and route frames to the corresponding models based on the policy to generate predictions. Our approach using separate models on ActivityNet (with ResNet-50) achieves an mAP of $74.9\%$ (an improvement of only $0.1\%$) while requiring $34.0$ GFLOPS and $115.6$MB of memory, in contrast to $28.1$ GFLOPS and $50.2$MB of memory with a single any-precision network. Similarly, use of separate models on Mini-Sports1M yields only $0.1\%$ improvement in mAP with $7.1\%$ more computation and $56.5\%$ of additional memory, compared to an any-precision network. This clearly shows the effectiveness of our any-precision network over individual quantized models in obtaining very competitive performance with less computation and memory.

\vspace{-1mm}
\section{Conclusion} \label{sec:conclusion}

In this paper, we introduce video instance-aware quantization that decides what precision should be used on a per frame basis for efficient video recognition. Specifically, we utilize a lightweight policy network to predict these decisions and train it in parallel with an any-precision recognition network with the goal of achieving both competitive accuracy and resource efficiency. Comprehensive experiments on four challenging and diverse datasets demonstrate the superiority of our approach over existing state-of-the-art methods.

\clearpage

{\small \noindent\textbf{Acknowledgements.} This work is also supported by the Intelligence Advanced Research Projects Activity (IARPA) via DOI/IBC contract number D17PC00341. The U.S. Government is authorized to reproduce and distribute reprints for Governmental purposes notwithstanding any copyright annotation thereon.}

{\small \noindent\textbf{Disclaimer.} The views and conclusions contained herein are those of the authors and should not be interpreted as necessarily representing the official policies or endorsements, either expressed or implied, of IARPA, DOI/IBC, or the U.S. Government.}

\newpage

\appendix

\section{Dataset Details} We evaluate our approach using four standard video recognition benchmark datasets, namely ActivityNet-v1.3~\cite{caba2015activitynet}, FCVID~\cite{jiang2017exploiting}, Mini-Sports1M~\cite{karpathy2014large} and Mini-Kinetics~\cite{carreira2017quo}. Below we provide more details on each of the dataset.

\vspace{1mm}
\noindent\textbf{ActivityNet.} We use the v1.3 split of ActivityNet dataset which consists of more than 648 hours of untrimmed videos from a total of 20K videos. Specifically, this dataset has 10,024 videos for training, 4926 videos for validation
and 5044 videos for testing with an average duration of 117 seconds. It contains 200 different daily activities such as: walking the dog, long
jump, and vacuuming floor. We use the training videos to train our network, and the validation set for testing as labels in the testing set are withheld by the authors. The dataset is publicly available to download at \url{http://activity-net.org/download.html}.

\vspace{1mm}
\noindent\textbf{FCVID.} Fudan-Columbia Video Dataset (FCVID) contains total 91,223 Web videos annotated manually according to 239 categories (45,611 videos for training and 45,612 videos for testing).
The categories cover a wide range of topics like social events, procedural events, objects, scenes, etc. that form in a hierarchy of 11 high-level groups (183 classes are related to events and 56 are objects, scenes, etc.). The total duration of FCVID is 4,232 hours with an average video duration of 167 seconds. The dataset is available to download at \url{http://bigvid.fudan.edu.cn/FCVID/}. 

\vspace{1mm}
\noindent\textbf{Mini-Sports1M.} Mini-Sports1M is a subset of Sports-1M~\cite{karpathy2014large} dataset with 1.1M videos of 487 different fine-grained sports. It is assembled by~\cite{gao2020listentolook} using videos of length 2-5 mins, and randomly sample 30 videos for each class for training, and 10 videos for each class for testing. The classes are arranged in a manually-curated taxonomy that contains internal nodes such as Aquatic Sports, Team Sports, Winter Sports, Ball Sports, etc, and generally becomes fine-grained by the leaf level. We obtain the training and testing splits from the authors of~\cite{gao2020listentolook} to perform our experiments. Both training and testing videos in this dataset are untrimmed. This dataset is available to download at \url{https://github.com/gtoderici/sports-1m-dataset}.

\vspace{1mm}
\noindent\textbf{Mini-Kinetics.} Kinetics-400 is a large-scale dataset containing 400 action classes and 240K training videos that are collected from YouTube. Since the full Kinetics dataset is quite large and the original version is no longer available from official site (about $\sim$15\% videos are missing), we use the Mini-Kinetics dataset that contains 121K videos for training and 10K videos for testing, with each video lasting 6-10 seconds. We use official training/validation splits of Mini-Kinetics released by authors~\cite{meng2020ar} in our experiments.

\section{Implementation Details}
\label{sec:impl}
\begin{table}
    \begin{center}
        \resizebox{\linewidth}{!}{
       \begin{tabular}{c|c|cc|cc|cc}
             \Xhline{3\arrayrulewidth} 
              \multirow{2}{*}{Arch.}  &  \multirow{2}{*}{$\alpha_{init}$} &  \multicolumn{2}{c|}{32-bit} &  \multicolumn{2}{c|}{4-bit} & \multicolumn{2}{c}{2-bit}   \\
              \cline{3-8}
              & & $\alpha_{lr}$ & $\alpha_{wd}$ &  $\alpha_{lr}$ & $\alpha_{wd}$  & $\alpha_{lr}$ & $\alpha_{wd}$  \\
             \Xhline{3\arrayrulewidth}  
             ResNet-18 & 4 & 0.01 & 5e-4 & 0.01 & 5e-4 & 0.01 & 5e-3 \\
            ResNet-50 & 2 & 0.1 & 5e-4 & 0.1 & 5e-4 & 0.01 & 6e-2 \\
            \Xhline{3\arrayrulewidth} 
        \end{tabular}
        } \vspace{-2mm}
    \caption{\small \textbf{Hyperparameters for training the any-precision recognition network}. We use separate sets of learning parameters (learning rate, weight decay) for clipping values of each precision.}
    \label{table:ap_hp}
    \end{center} \vspace{-4mm}
\end{table}
\begin{table}
    \begin{center}
        \resizebox{0.6\linewidth}{!}{
       \begin{tabular}{c|c c c}
             \Xhline{3\arrayrulewidth} 
              Dataset &  $w_1$ & $w_2$ & $ w_3$ \\
             \Xhline{3\arrayrulewidth}  
            ActivityNet & 0.21 & 0.5 & 0.1 \\
            FCVID & 0.11 & 1.0 & 0.1 \\
            Mini-Sports1M &  0.21 & 0.5 & 0.1 \\
            Mini-Kinetics &  0.21 & 0.3 & 0.1\\
            \Xhline{3\arrayrulewidth} 
        \end{tabular}
        } \vspace{-2mm}
    \caption{\small \textbf{Hyperparameters to train the policy network}.}
    \label{table:adap_hp}
    \end{center} \vspace{-5mm}
\end{table}

In this section, we provide more details regarding the implementation. 
We train the any-precision recognition network from the full-precision recognition network pretrained on the same dataset for 100 epochs. Then we optimize the policy network accompanied with the well-trained (frozen) any-precision recognition network for 50 epochs and the policy network is initialized with the weight pretrained on the same dataset as well.
For our experiments, we use 12 NVIDIA Tesla V100 GPUs for training the any-precision recognition network and 6 GPUSs for training the policy network. All our models were implemented and trained via PyTorch. In Table~\ref{table:ap_hp} and \ref{table:adap_hp},  we provide the initial value ($\alpha_{init}$), learning rate ($\alpha_{lr}$) and weight decay ($\alpha_{wd}$) for each precision to train the any-precision recognition network, as well as hyperparameters $w_1$, $w_2$ and $w_3$ (in Eq. (13) in the main paper) to train the policy network. 
The data augmentations in our approach are based on the practices in~\cite{wang2018non}. We first randomly resize the shorter side of an image to a range of [256, 320) while keeping aspect ratio and then randomly crop a $224\times224$ region and normalize it with the ImageNet's mean and standard deviation to form the input ($16\times224\times224$). The training time depends on the size of datasets and the task. We will make our code publicly available after the acceptance.

\section{Additional Ablation Studies}
\vspace{1mm}
\noindent\textbf{Effectiveness of LSTM.}
We investigate the effectiveness of LSTM for modeling video causality in the policy network by comparing with a variant of \ours without LSTM (see Table~\ref{table:lstm}). On ActivityNet and Mini-Sports1M datasets, the variant without LSTM yields $0.7\%$ and $0.3\%$ lower mAP with similar GFLOPs than \ours respectively. 
This demonstrates that LSTM is critical for good performance as it makes the policy network aware of all useful information seen so far by aggregating the sequence history.

\begin{table}
    \begin{center}
        \resizebox{0.6\linewidth}{!}{
       \begin{tabular}{c|cc}
             \Xhline{3\arrayrulewidth} 
             Model &  mAP (\%) & GFLOPs    \\
             \Xhline{3\arrayrulewidth}  
             \multicolumn{3}{c}{ActivityNet} \\
             \hline
            No LSTM & 74.1 & 28.8 \\
             LSTM & \textbf{74.8} & 28.1  \\
             \hline
             \multicolumn{3}{c}{Mini-Kinetics} \\
             \hline
             No LSTM & 46.1 & 	26.4 \\
            LSTM & \textbf{46.4} & 26.8  \\
            \Xhline{3\arrayrulewidth} 
        \end{tabular}
        } \vspace{-2mm}
    \caption{\small \textbf{Effect of LSTM on ActivityNet and Mini-Sports1M}.}
    \label{table:lstm}
    \end{center} \vspace{-5mm}
\end{table}

\vspace{1mm}
\noindent\textbf{Effect of Different Losses.}
Similar to Table 5 of the main paper, we further ablate different losses on Mini-Sports1M (see Table~\ref{table:ablation_loss_sports1m}) and observe that without knowledge transfer from a pretrained full-precision model, our method only achieves $44.6\%$ with similar amount of GFLOPs. It once again demonstrates the importance of using the full-precision model as the teacher for effective training of lower precisions. When training without efficiency loss (by setting $\mathcal{L}_e=0$), it achieves $46.6\%$ mAP ($0.2\%$ improvement) but with $118\%$ more FLOPs. Furthermore, $\mathcal{L}_b$ and $\mathcal{L}_d$ both improve the performance with similar computational cost.
\begin{table}
    \begin{center}
        \resizebox{0.8\linewidth}{!}{
       \begin{tabular}{ccccc|cc}
             \Xhline{3\arrayrulewidth} 
              $\mathcal{L}_{ce}$ & $\mathcal{L}_{kd}$ & $\mathcal{L}_{e}$ & $\mathcal{L}_{b}$ & $\mathcal{L}_{d}$ &  mAP (\%) & GFLOPs   \\
             \Xhline{3\arrayrulewidth}  
           \checkmark & &  \checkmark  &  \checkmark  &   \checkmark &  44.6 & 26.5    \\
           \checkmark & \checkmark & & & & 46.6  & 58.5  \\
        \checkmark & \checkmark &\checkmark & & & 46.3 & 28.5 \\
            \checkmark & \checkmark &\checkmark &\checkmark & &  46.2 & 26.9 \\
            \checkmark & \checkmark &\checkmark &\checkmark & \checkmark &   \textbf{46.4} & 26.8 \\
            \Xhline{3\arrayrulewidth} 
        \end{tabular}
        } \vspace{-2mm}
          \caption{\small \textbf{Effect of different losses on Mini-Sports1M}.}
        \label{table:ablation_loss_sports1m}
    \end{center} \vspace{-3mm}
\end{table}

\vspace{1mm}
\noindent\textbf{Effect of Decision Space.}
Similar to Table 6 in main paper, we show the effect of decision space $\Omega$ on Mini-Sports1M (see Table~\ref{table:different_b_sports1m}). We adjust the training loss to keep their GFLOPS at the same level and we only compare the differences in recognition performances. Only skipping frames yields $43.9\%$ in mAP ($0.5\%$ lower than $\Omega = \{32, 4, 2, 0\}$). Among all the alternatives, the best strategy is to set $\Omega = \{32, 4, 2, 0\}$ for achieving top performance of $46.4\%$ in mAP with $26.8$ GFLOPs.

\begin{table}
    \begin{center}
        \resizebox{0.8\linewidth}{!}{
       \begin{tabular}{c|cc}
             \Xhline{3\arrayrulewidth} 
              Decision Space $\Omega$ &  mAP (\%) & GFLOPs    \\
             \Xhline{3\arrayrulewidth}  
            \{32, 0\} &  43.9 & 	28.7 \\
             \{32, 4, 2\} & 46.1 & 	29.3 \\
            \{32, 4, 0\} & 43.9 & 	33.5 \\
             \{32, 2, 0\} &  46.0 & 32.9  \\
            \{32, 4, 2, 0\} & \textbf{46.4} &\textbf{ 26.8 } \\
            \Xhline{3\arrayrulewidth} 
        \end{tabular}
        } \vspace{-2mm}
    \caption{\small \textbf{Effect of different decision space on Mini-Sports1M}.}
    \label{table:different_b_sports1m}
    \end{center} \vspace{-5mm}
\end{table}

\section{Qualitative Results}
\label{sec:qual}

\begin{figure*}
\begin{center}
     \includegraphics[width=0.82\linewidth]{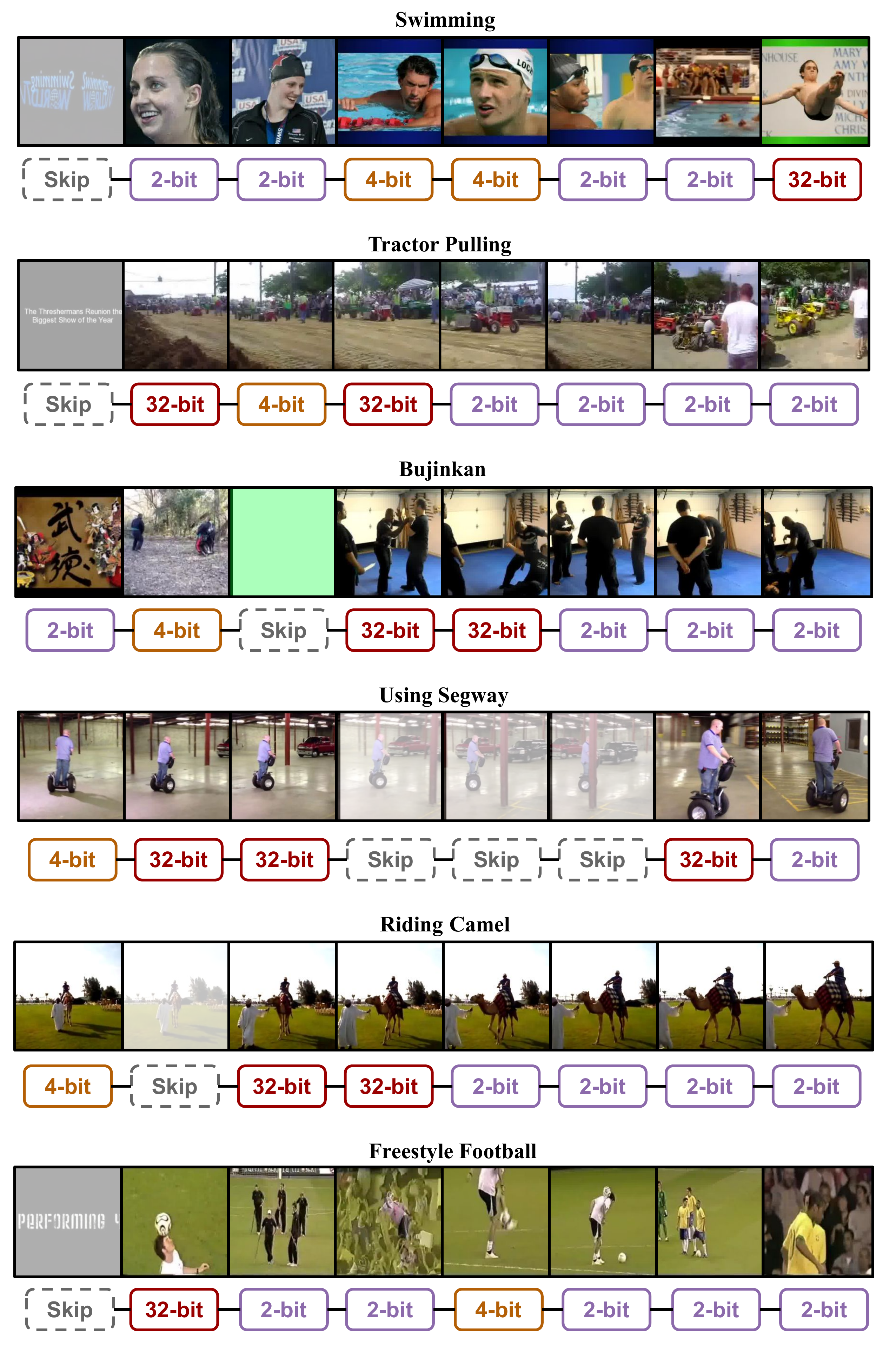}
\end{center} \vspace{-6mm}
   \caption{\small \textbf{Qualitative examples.} Our proposed approach \ours processes more informative frames with high precision and less informative ones with lower precision or skip them when irrelevant, for efficient video recognition. Best viewed in color.  
   }
   \vspace{-5pt}
   \label{fig:qualitative_results_supp}
\end{figure*}

In this section, we provide additional qualitative examples to visualize the learnt policy (see Figure~\ref{fig:qualitative_results_supp}).  Videos are uniformly sampled in 8 frames. \ours processes most informative frames with 32-bit precision while it skips or uses lower precision for the less informative frames without sacrificing accuracy (see top 4 examples in Figure~\ref{fig:qualitative_results_supp}: \enquote{Swimming}, \enquote{Tractor Pulling}, \enquote{Bujinkan} and \enquote{Using Segway}). Moreover, it uses 2-bit precision instead of 32-bit precision (see bottom 2 examples in Figure~\ref{fig:qualitative_results_supp}: \enquote{Riding Camel} and \enquote{Freestyle Football}) after being confident about the action. 


{\small
\begin{thebibliography}{10}\itemsep=-1pt

\bibitem{bengio2015conditional}
Emmanuel Bengio, Pierre-Luc Bacon, Joelle Pineau, and Doina Precup.
\newblock Conditional computation in neural networks for faster models.
\newblock {\em arXiv preprint arXiv:1511.06297}, 2015.

\bibitem{bengio2013estimating}
Yoshua Bengio, Nicholas L{\'e}onard, and Aaron Courville.
\newblock Estimating or propagating gradients through stochastic neurons for
  conditional computation.
\newblock {\em arXiv preprint arXiv:1308.3432}, 2013.

\bibitem{caba2015activitynet}
Fabian Caba~Heilbron, Victor Escorcia, Bernard Ghanem, and Juan Carlos~Niebles.
\newblock Activitynet: A large-scale video benchmark for human activity
  understanding.
\newblock In {\em Proceedings of the ieee conference on computer vision and
  pattern recognition}, pages 961--970, 2015.

\bibitem{cai2020rethinking}
Zhaowei Cai and Nuno Vasconcelos.
\newblock Rethinking differentiable search for mixed-precision neural networks.
\newblock In {\em Proceedings of the IEEE/CVF Conference on Computer Vision and
  Pattern Recognition}, pages 2349--2358, 2020.

\bibitem{carreira2017quo}
Joao Carreira and Andrew Zisserman.
\newblock Quo vadis, action recognition? a new model and the kinetics dataset.
\newblock In {\em proceedings of the IEEE Conference on Computer Vision and
  Pattern Recognition}, pages 6299--6308, 2017.

\bibitem{chen2020deep}
Chun-Fu Chen, Rameswar Panda, Kandan Ramakrishnan, Rogerio Feris, John Cohn,
  Aude Oliva, and Quanfu Fan.
\newblock Deep analysis of cnn-based spatio-temporal representations for action
  recognition.
\newblock {\em arXiv preprint arXiv:2010.11757}, 2020.

\bibitem{chen2019you}
Zhourong Chen, Yang Li, Samy Bengio, and Si Si.
\newblock You look twice: Gaternet for dynamic filter selection in cnns.
\newblock In {\em Proceedings of the IEEE Conference on Computer Vision and
  Pattern Recognition}, pages 9172--9180, 2019.

\bibitem{choi2018pact}
Jungwook Choi, Zhuo Wang, Swagath Venkataramani, Pierce I-Jen Chuang,
  Vijayalakshmi Srinivasan, and Kailash Gopalakrishnan.
\newblock Pact: Parameterized clipping activation for quantized neural
  networks.
\newblock {\em arXiv preprint arXiv:1805.06085}, 2018.

\bibitem{fan2018watching}
Hehe Fan, Zhongwen Xu, Linchao Zhu, Chenggang Yan, Jianjun Ge, and Yi Yang.
\newblock Watching a small portion could be as good as watching all: Towards
  efficient video classification.
\newblock In {\em IJCAI International Joint Conference on Artificial
  Intelligence}, 2018.

\bibitem{fan2019more}
Quanfu Fan, Chun-Fu~Richard Chen, Hilde Kuehne, Marco Pistoia, and David Cox.
\newblock More is less: Learning efficient video representations by big-little
  network and depthwise temporal aggregation.
\newblock In {\em Advances in Neural Information Processing Systems}, pages
  2261--2270, 2019.

\bibitem{feichtenhofer2020x3d}
Christoph Feichtenhofer.
\newblock X3d: Expanding architectures for efficient video recognition.
\newblock {\em arXiv preprint arXiv:2004.04730}, 2020.

\bibitem{feichtenhofer2019slowfast}
Christoph Feichtenhofer, Haoqi Fan, Jitendra Malik, and Kaiming He.
\newblock Slowfast networks for video recognition.
\newblock In {\em Proceedings of the IEEE International Conference on Computer
  Vision}, pages 6202--6211, 2019.

\bibitem{figurnov2017spatially}
Michael Figurnov, Maxwell~D Collins, Yukun Zhu, Li Zhang, Jonathan Huang,
  Dmitry Vetrov, and Ruslan Salakhutdinov.
\newblock Spatially adaptive computation time for residual networks.
\newblock In {\em Proceedings of the IEEE Conference on Computer Vision and
  Pattern Recognition}, pages 1039--1048, 2017.

\bibitem{gao2020listentolook}
{Gao, Ruohan and Oh, Tae-Hyun, and Grauman, Kristen and Torresani, Lorenzo}.
\newblock Listen to look: Action recognition by previewing audio.
\newblock In {\em Proceedings of the IEEE Conference on Computer Vision and
  Pattern Recognition (CVPR)}, 2020.

\bibitem{guo2019spottune}
Yunhui Guo, Honghui Shi, Abhishek Kumar, Kristen Grauman, Tajana Rosing, and
  Rogerio Feris.
\newblock Spottune: transfer learning through adaptive fine-tuning.
\newblock In {\em Proceedings of the IEEE Conference on Computer Vision and
  Pattern Recognition}, pages 4805--4814, 2019.

\bibitem{han2015deep}
Song Han, Huizi Mao, and William~J Dally.
\newblock Deep compression: Compressing deep neural networks with pruning,
  trained quantization and huffman coding.
\newblock {\em arXiv preprint arXiv:1510.00149}, 2015.

\bibitem{hara2018can}
Kensho Hara, Hirokatsu Kataoka, and Yutaka Satoh.
\newblock Can spatiotemporal 3d cnns retrace the history of 2d cnns and
  imagenet?
\newblock In {\em Proceedings of the IEEE conference on Computer Vision and
  Pattern Recognition}, pages 6546--6555, 2018.

\bibitem{he2016deep}
Kaiming He, Xiangyu Zhang, Shaoqing Ren, and Jian Sun.
\newblock Deep residual learning for image recognition.
\newblock In {\em Proceedings of the IEEE conference on computer vision and
  pattern recognition}, pages 770--778, 2016.

\bibitem{hinton2015distilling}
Geoffrey Hinton, Oriol Vinyals, and Jeff Dean.
\newblock Distilling the knowledge in a neural network.
\newblock {\em arXiv preprint arXiv:1503.02531}, 2015.

\bibitem{hua2019channel}
Weizhe Hua, Yuan Zhou, Christopher~M De~Sa, Zhiru Zhang, and G~Edward Suh.
\newblock Channel gating neural networks.
\newblock In {\em Advances in Neural Information Processing Systems}, pages
  1884--1894, 2019.

\bibitem{huang2017multi}
Gao Huang, Danlu Chen, Tianhong Li, Felix Wu, Laurens van~der Maaten, and
  Kilian~Q Weinberger.
\newblock Multi-scale dense networks for resource efficient image
  classification.
\newblock {\em arXiv preprint arXiv:1703.09844}, 2017.

\bibitem{hubara2016binarized}
Itay Hubara, Matthieu Courbariaux, Daniel Soudry, Ran El-Yaniv, and Yoshua
  Bengio.
\newblock Binarized neural networks.
\newblock In {\em Advances in neural information processing systems}, pages
  4107--4115, 2016.

\bibitem{hussein2020timegate}
Noureldien Hussein, Mihir Jain, and Babak~Ehteshami Bejnordi.
\newblock Timegate: Conditional gating of segments in long-range activities.
\newblock {\em arXiv preprint arXiv:2004.01808}, 2020.

\bibitem{jang2016categorical}
Eric Jang, Shixiang Gu, and Ben Poole.
\newblock Categorical reparameterization with gumbel-softmax.
\newblock In {\em International Conference on Learning Representations}, 2017.

\bibitem{jiang2017exploiting}
Yu-Gang Jiang, Zuxuan Wu, Jun Wang, Xiangyang Xue, and Shih-Fu Chang.
\newblock Exploiting feature and class relationships in video categorization
  with regularized deep neural networks.
\newblock {\em IEEE transactions on pattern analysis and machine intelligence},
  40(2):352--364, 2017.

\bibitem{jie2019anytime}
Zequn Jie, Peng Sun, Xin Li, Jiashi Feng, and Wei Liu.
\newblock Anytime recognition with routing convolutional networks.
\newblock {\em IEEE transactions on pattern analysis and machine intelligence},
  2019.

\bibitem{jin2020adabits}
Qing Jin, Linjie Yang, and Zhenyu Liao.
\newblock Adabits: Neural network quantization with adaptive bit-widths.
\newblock In {\em Proceedings of the IEEE/CVF Conference on Computer Vision and
  Pattern Recognition}, pages 2146--2156, 2020.

\bibitem{karpathy2014large}
Andrej Karpathy, George Toderici, Sanketh Shetty, Thomas Leung, Rahul
  Sukthankar, and Li Fei-Fei.
\newblock Large-scale video classification with convolutional neural networks.
\newblock In {\em Proceedings of the IEEE conference on Computer Vision and
  Pattern Recognition}, pages 1725--1732, 2014.

\bibitem{kim2019qkd}
Jangho Kim, Yash Bhalgat, Jinwon Lee, Chirag Patel, and Nojun Kwak.
\newblock Qkd: Quantization-aware knowledge distillation.
\newblock {\em arXiv preprint arXiv:1911.12491}, 2019.

\bibitem{korbar2019scsampler}
Bruno Korbar, Du Tran, and Lorenzo Torresani.
\newblock Scsampler: Sampling salient clips from video for efficient action
  recognition.
\newblock In {\em Proceedings of the IEEE International Conference on Computer
  Vision}, pages 6232--6242, 2019.

\bibitem{leng2018extremely}
Cong Leng, Zesheng Dou, Hao Li, Shenghuo Zhu, and Rong Jin.
\newblock Extremely low bit neural network: Squeeze the last bit out with admm.
\newblock In {\em Proceedings of the AAAI Conference on Artificial
  Intelligence}, volume~32, 2018.

\bibitem{lin2019tsm}
Ji Lin, Chuang Gan, and Song Han.
\newblock Tsm: Temporal shift module for efficient video understanding.
\newblock In {\em Proceedings of the IEEE International Conference on Computer
  Vision}, pages 7083--7093, 2019.

\bibitem{mcgill2017deciding}
Mason McGill and Pietro Perona.
\newblock Deciding how to decide: Dynamic routing in artificial neural
  networks.
\newblock In {\em Proceedings of the 34th International Conference on Machine
  Learning-Volume 70}, pages 2363--2372, 2017.

\bibitem{meng2020ar}
Yue Meng, Chung-Ching Lin, Rameswar Panda, Prasanna Sattigeri, Leonid
  Karlinsky, Aude Oliva, Kate Saenko, and Rogerio Feris.
\newblock Ar-net: Adaptive frame resolution for efficient action recognition.
\newblock In {\em ECCV}, 2020.

\bibitem{meng2021adafuse}
Yue Meng, Rameswar Panda, Chung-Ching Lin, Prasanna Sattigeri, Leonid
  Karlinsky, Kate Saenko, Aude Oliva, and Rogerio Feris.
\newblock Adafuse: Adaptive temporal fusion network for efficient action
  recognition.
\newblock In {\em International Conference on Learning Representations}, 2021.

\bibitem{monfort2018moments}
Mathew Monfort, Alex Andonian, Bolei Zhou, Kandan Ramakrishnan, Sarah~Adel
  Bargal, Tom Yan, Lisa Brown, Quanfu Fan, Dan Gutfruend, Carl Vondrick, et~al.
\newblock Moments in time dataset: one million videos for event understanding.
\newblock {\em arXiv preprint arXiv:1801.03150}, 2018.

\bibitem{najibi2019autofocus}
Mahyar Najibi, Bharat Singh, and Larry~S Davis.
\newblock Autofocus: Efficient multi-scale inference.
\newblock In {\em Proceedings of the IEEE International Conference on Computer
  Vision}, pages 9745--9755, 2019.

\bibitem{pan2021vared2}
Bowen Pan, Rameswar Panda, Camilo~Luciano Fosco, Chung-Ching Lin, Alex~J
  Andonian, Yue Meng, Kate Saenko, Aude Oliva, and Rogerio Feris.
\newblock Va-red$^2$: Video adaptive redundancy reduction.
\newblock In {\em International Conference on Learning Representations}, 2021.

\bibitem{park2017weighted}
Eunhyeok Park, Junwhan Ahn, and Sungjoo Yoo.
\newblock Weighted-entropy-based quantization for deep neural networks.
\newblock In {\em Proceedings of the IEEE Conference on Computer Vision and
  Pattern Recognition}, pages 5456--5464, 2017.

\bibitem{phan2020binarizing}
Hai Phan, Zechun Liu, Dang Huynh, Marios Savvides, Kwang-Ting Cheng, and
  Zhiqiang Shen.
\newblock Binarizing mobilenet via evolution-based searching.
\newblock In {\em Proceedings of the IEEE/CVF Conference on Computer Vision and
  Pattern Recognition}, pages 13420--13429, 2020.

\bibitem{piergiovanni2019tiny}
AJ Piergiovanni, Anelia Angelova, and Michael~S Ryoo.
\newblock Tiny video networks.
\newblock {\em arXiv preprint arXiv:1910.06961}, 2019.

\bibitem{rastegari2016xnor}
Mohammad Rastegari, Vicente Ordonez, Joseph Redmon, and Ali Farhadi.
\newblock Xnor-net: Imagenet classification using binary convolutional neural
  networks.
\newblock In {\em European conference on computer vision}, pages 525--542.
  Springer, 2016.

\bibitem{sandler2018mobilenetv2}
Mark Sandler, Andrew Howard, Menglong Zhu, Andrey Zhmoginov, and Liang-Chieh
  Chen.
\newblock Mobilenetv2: Inverted residuals and linear bottlenecks.
\newblock In {\em Proceedings of the IEEE conference on computer vision and
  pattern recognition}, pages 4510--4520, 2018.

\bibitem{shen2020once}
Mingzhu Shen, Feng Liang, Chuming Li, Chen Lin, Ming Sun, Junjie Yan, and Wanli
  Ouyang.
\newblock Once quantized for all: Progressively searching for quantized
  efficient models.
\newblock {\em arXiv preprint arXiv:2010.04354}, 2020.

\bibitem{Simonyan14TwoStream}
Karen Simonyan and Andrew Zisserman.
\newblock Two-stream convolutional networks for action recognition in videos.
\newblock In {\em Neural Information Processing System (NIPS)}, 2014.

\bibitem{sudhakaran2020gate}
Swathikiran Sudhakaran, Sergio Escalera, and Oswald Lanz.
\newblock Gate-shift networks for video action recognition.
\newblock In {\em Proceedings of the IEEE/CVF Conference on Computer Vision and
  Pattern Recognition}, pages 1102--1111, 2020.

\bibitem{sun2020adashare}
Ximeng Sun, Rameswar Panda, Rogerio Feris, and Kate Saenko.
\newblock Adashare: Learning what to share for efficient deep multi-task
  learning.
\newblock {\em Advances in Neural Information Processing Systems}, 33, 2020.

\bibitem{tran2015learning}
Du Tran, Lubomir Bourdev, Rob Fergus, Lorenzo Torresani, and Manohar Paluri.
\newblock Learning spatiotemporal features with 3d convolutional networks.
\newblock In {\em Proceedings of the IEEE international conference on computer
  vision}, pages 4489--4497, 2015.

\bibitem{tran2019video}
Du Tran, Heng Wang, Lorenzo Torresani, and Matt Feiszli.
\newblock Video classification with channel-separated convolutional networks.
\newblock In {\em Proceedings of the IEEE International Conference on Computer
  Vision}, pages 5552--5561, 2019.

\bibitem{veit2018convolutional}
Andreas Veit and Serge Belongie.
\newblock Convolutional networks with adaptive inference graphs.
\newblock In {\em Proceedings of the European Conference on Computer Vision
  (ECCV)}, pages 3--18, 2018.

\bibitem{wang2019haq}
Kuan Wang, Zhijian Liu, Yujun Lin, Ji Lin, and Song Han.
\newblock Haq: Hardware-aware automated quantization with mixed precision.
\newblock In {\em Proceedings of the IEEE conference on computer vision and
  pattern recognition}, pages 8612--8620, 2019.

\bibitem{wang2016temporal}
Limin Wang, Yuanjun Xiong, Zhe Wang, Yu Qiao, Dahua Lin, Xiaoou Tang, and Luc
  Van~Gool.
\newblock Temporal segment networks: Towards good practices for deep action
  recognition.
\newblock In {\em European conference on computer vision}, pages 20--36.
  Springer, 2016.

\bibitem{wang2018non}
Xiaolong Wang, Ross Girshick, Abhinav Gupta, and Kaiming He.
\newblock Non-local neural networks.
\newblock In {\em Proceedings of the IEEE conference on computer vision and
  pattern recognition}, pages 7794--7803, 2018.

\bibitem{wang2018skipnet}
Xin Wang, Fisher Yu, Zi-Yi Dou, Trevor Darrell, and Joseph~E Gonzalez.
\newblock Skipnet: Learning dynamic routing in convolutional networks.
\newblock In {\em Proceedings of the European Conference on Computer Vision
  (ECCV)}, pages 409--424, 2018.

\bibitem{wang2019learning}
Ziwei Wang, Jiwen Lu, Chenxin Tao, Jie Zhou, and Qi Tian.
\newblock Learning channel-wise interactions for binary convolutional neural
  networks.
\newblock In {\em Proceedings of the IEEE/CVF Conference on Computer Vision and
  Pattern Recognition}, pages 568--577, 2019.

\bibitem{williams1992simple}
Ronald~J Williams.
\newblock Simple statistical gradient-following algorithms for connectionist
  reinforcement learning.
\newblock {\em Machine learning}, 8(3-4):229--256, 1992.

\bibitem{wu2018mixed}
Bichen Wu, Yanghan Wang, Peizhao Zhang, Yuandong Tian, Peter Vajda, and Kurt
  Keutzer.
\newblock Mixed precision quantization of convnets via differentiable neural
  architecture search.
\newblock {\em arXiv preprint arXiv:1812.00090}, 2018.

\bibitem{wu2019multi}
Wenhao Wu, Dongliang He, Xiao Tan, Shifeng Chen, and Shilei Wen.
\newblock Multi-agent reinforcement learning based frame sampling for effective
  untrimmed video recognition.
\newblock In {\em Proceedings of the IEEE International Conference on Computer
  Vision}, pages 6222--6231, 2019.

\bibitem{wu2018blockdrop}
Zuxuan Wu, Tushar Nagarajan, Abhishek Kumar, Steven Rennie, Larry~S Davis,
  Kristen Grauman, and Rogerio Feris.
\newblock Blockdrop: Dynamic inference paths in residual networks.
\newblock In {\em Proceedings of the IEEE Conference on Computer Vision and
  Pattern Recognition}, pages 8817--8826, 2018.

\bibitem{wu2019liteeval}
Zuxuan Wu, Caiming Xiong, Yu-Gang Jiang, and Larry~S Davis.
\newblock Liteeval: A coarse-to-fine framework for resource efficient video
  recognition.
\newblock In {\em Advances in Neural Information Processing Systems}, pages
  7778--7787, 2019.

\bibitem{wu2019adaframe}
Zuxuan Wu, Caiming Xiong, Chih-Yao Ma, Richard Socher, and Larry~S Davis.
\newblock Adaframe: Adaptive frame selection for fast video recognition.
\newblock In {\em Proceedings of the IEEE Conference on Computer Vision and
  Pattern Recognition}, pages 1278--1287, 2019.

\bibitem{yang2020resolution}
Le Yang, Yizeng Han, Xi Chen, Shiji Song, Jifeng Dai, and Gao Huang.
\newblock Resolution adaptive networks for efficient inference.
\newblock In {\em Proceedings of the IEEE/CVF Conference on Computer Vision and
  Pattern Recognition}, pages 2369--2378, 2020.

\bibitem{yang2020mutualnet}
Taojiannan Yang, Sijie Zhu, Chen Chen, Shen Yan, Mi Zhang, and Andrew Willis.
\newblock Mutualnet: Adaptive convnet via mutual learning from network width
  and resolution.
\newblock In {\em European Conference on Computer Vision (ECCV)}, 2020.

\bibitem{yeung2016end}
Serena Yeung, Olga Russakovsky, Greg Mori, and Li Fei-Fei.
\newblock End-to-end learning of action detection from frame glimpses in
  videos.
\newblock In {\em Proceedings of the IEEE Conference on Computer Vision and
  Pattern Recognition}, pages 2678--2687, 2016.

\bibitem{yu2019any}
Haichao Yu, Haoxiang Li, Honghui Shi, Thomas~S Huang, and Gang Hua.
\newblock Any-precision deep neural networks.
\newblock {\em arXiv preprint arXiv:1911.07346}, 2019.

\bibitem{yu2018slimmable}
Jiahui Yu, Linjie Yang, Ning Xu, Jianchao Yang, and Thomas Huang.
\newblock Slimmable neural networks.
\newblock {\em arXiv preprint arXiv:1812.08928}, 2018.

\bibitem{zhang2018lq}
Dongqing Zhang, Jiaolong Yang, Dongqiangzi Ye, and Gang Hua.
\newblock Lq-nets: Learned quantization for highly accurate and compact deep
  neural networks.
\newblock In {\em Proceedings of the European conference on computer vision
  (ECCV)}, pages 365--382, 2018.

\bibitem{zhou2018temporal}
Bolei Zhou, Alex Andonian, Aude Oliva, and Antonio Torralba.
\newblock Temporal relational reasoning in videos.
\newblock In {\em Proceedings of the European Conference on Computer Vision
  (ECCV)}, pages 803--818, 2018.

\bibitem{zhou2016dorefa}
Shuchang Zhou, Yuxin Wu, Zekun Ni, Xinyu Zhou, He Wen, and Yuheng Zou.
\newblock Dorefa-net: Training low bitwidth convolutional neural networks with
  low bitwidth gradients.
\newblock {\em arXiv preprint arXiv:1606.06160}, 2016.

\bibitem{zhuang2020training}
Bohan Zhuang, Lingqiao Liu, Mingkui Tan, Chunhua Shen, and Ian Reid.
\newblock Training quantized neural networks with a full-precision auxiliary
  module.
\newblock In {\em Proceedings of the IEEE/CVF Conference on Computer Vision and
  Pattern Recognition}, pages 1488--1497, 2020.

\bibitem{zhuang2018towards}
Bohan Zhuang, Chunhua Shen, Mingkui Tan, Lingqiao Liu, and Ian Reid.
\newblock Towards effective low-bitwidth convolutional neural networks.
\newblock In {\em Proceedings of the IEEE conference on computer vision and
  pattern recognition}, pages 7920--7928, 2018.

\end{thebibliography}

}
\end{document}